\documentclass[11pt]{article}

\usepackage[a4paper,margin=1in]{geometry}
\usepackage{authblk}
\usepackage{graphicx}
\usepackage{booktabs}
\usepackage{amsmath,amssymb}
\usepackage{natbib}
\usepackage{hyperref}
\usepackage{orcidlink} 

\usepackage{url,hyperref,lineno,microtype,subcaption}
\usepackage[onehalfspacing]{setspace}
\usepackage{booktabs}

\author[1]{Morten Roed Frederiksen}
\author[2]{Kasper Støy}
\author[3]{Maja Matarić}
\affil[1, 2]{Data Systems and Robotics, IT-University of Copenhagen, Denmark}
\affil[3]{Interaction Lab, University of Southern California, Los Angeles, California, U.S.}

\date{}

\begin{document}
\onecolumn

\title{Inducing Calmness With Pocket-Sized Robotics: Reducing Movement and Heart Rate in Children through Hand-Held Tactile Interactions}


\maketitle

\begin{abstract}

Periods of heightened arousal or restlessness can interfere with children’s ability to focus, self-regulation, and physically calm. Technologies that encourage embodied self-regulation through tactile interaction may provide a simple and accessible means of promoting calmness. This paper investigates how interaction with a pocket-sized tactile device influences physiological and behavioral markers of calmness in typically developing children. Building on prior work examining heart rate modulation, we present new findings on how tactile interaction affects full-body movement and postural stability. We employ a device that engages children through a hand-held rhythmic vibration-matching game, designed to focus attention and encourage stillness. Eighteen children participated in a within-subjects study that involved two conditions: with and without tactile interaction with a hand-held device, while having their heart rate and body movement recorded. Results show that the tactile game interaction reduced physiological arousal (heart rate decreased by 3.37 bpm $p = 0.019$) and physical restlessness (overall movement decreased by 37.6\% $p = 0.02$), with attention-related body regions showing the greatest change toward stillness (44.8\% reduction in movement).These findings demonstrate that brief tactile game-like engagement with a hand-held device can down-regulate physiological activation, promoting the calm and focused states toward sustained attention and behavior regulation.
\end{abstract}

\section{Introduction}\label{sec:intro}

Supporting children’s ability to self-regulate, control impulses, and be able to be physically calm is central to healthy cognitive and emotional development \citep{Widiger2022TheDA, Nilfyr2025ThrivingCE}. In everyday contexts such as classrooms and home environments, children experience states of heightened arousal, restlessness, or difficulty maintaining focused attention \citep{Ciucci2016EmotionalAA, Nkrumah2015EffectOC}. These conditions can manifest through physiological indicators such as elevated heart rate, and behavioral expressions such as fidgeting or excessive movement \citep{Komissarova2024HeartRV, Dogadkina2022HeartRV, Bang2017ACO}. Finding accessible and engaging ways to help children regulate these responses remains an ongoing challenge.

Traditional approaches to fostering calmness in children—such as mindfulness exercises and breathing techniques—can be effective but often require repeated instruction, structured practice, and adult facilitation \citep{Li2021TheRA, Jain2021ImpactOY}. As a complement to such methods, interactive technologies that provide immediate or delayed sensory feedback may offer a simple and playful pathway toward increased attention and self-regulation \citep{Ayearst2023AnOS, Basu2022AssessmentOA}. Tactile engagement, in particular, can redirect attention to the body and promote stillness through rhythmic or patterned touch \citep{BincyO2019EffectivenessOT, Davies2019TheEF}. Intelligent technology has the potential to address key gaps in existing interventions by delivering immediate forms of support that require little or no prior training, potentially making it more accessible to self-regulate emotions in difficult scenarios \citep{Rasouli2022PotentialAO}. This research explores whether a small hand-held tactile device can influence physiological and behavioral markers of calmness in typically developing children. Building on our earlier study showing reductions in heart rate during tactile interaction with a pocket device, here we extend the analysis to include movement tracking to assess changes in general restlessness and postural stability \citep{Frederiksen2024TactileCL}. We investigate how a brief rhythm-based tactile game affects both physiological arousal, defined as the level of physiological activation that dictates motor readiness and the intensity of physical movement, providing a quantitative assessment of embodied calmness \cite{Erhlich63,Duffy1957ThePS,Smith2024ArousalMN}. By investigating how tactile games induce calmness in a healthy population, we establish a normative baseline for future applications in therapeutic and other sensitive environments.

\subsection{Tactile Interaction and Self-Regulation}
Previous research in child–technology interaction has shown that gentle touch, rhythmic feedback, and simple embodied tasks can influence attention and emotion regulation \citep{BincyO2019EffectivenessOT, Ren2025TouchedBC}. Tactile engagement provides a form of sensory grounding that may assist children in maintaining focus and reducing unnecessary movement \citep{Cha2023ExploringTE}. Additionally, related work in sensory integration and self-regulation has demonstrated that tactile input can stabilize posture and modulate physiological states by guiding awareness toward the body \citep{Frederiksen2024TactileCL, Wang2021OrganizationAU, Sabinson2021PlantHumanEB}. Similarly, socially assistive and therapeutic robots have shown promise as means of encouraging calmness, focus, and engagement through haptic or rhythmic cues \citep{Kimura2023DevelopmentOA, Sefidgar2016DesignAE, Ma2021RhythmicBM}.

Our work examines these mechanisms using a pocket-sized tactile device (shown in Figure \ref{affecta_annotated}) that invites the user to engage in rhythmic matching through vibration and button presses. As a small form factor ubiquitous technology in the shape of a small robot, the device is conceptualized as a tool for embodied self-regulation. It is developed to explore how tactile feedback and attention coupling can influence physiological calmness. By studying changes in users' heart rate and physical movement, we aim to understand how short-term tactile engagement with the device can promote a calmer, more focused bodily state in children.

\subsection{Extending to Embodied Self-Control: A Hypothesis About Movement Modulation}
While much of prior research on self-regulation has focused on internal physiological responses such as heart rate or breathing, children’s ability to remain calm also depends on external behavioral control \citep{Nilfyr2025ThrivingCE, Eisenberg2010EmotionrelatedSA}. Small, seemingly spontaneous body movements—such as fidgeting, shifting weight, or moving the arms and legs---often reflect fluctuating levels of arousal \citep{Rosenbaum2021BodyMA, Gerdemann2025BodyPA}. These movements can interfere with attention and signal reduced calmness \citep{Raghunathan2022MoreTM}. The concept of embodied self-regulation proposes that calmness emerges not only from internal physiological balance but also from a stable and composed physical posture \citep{Karlesky2016UnderstandingFW}.

In this study, we investigated whether the tactile interaction that lowers heart rate also reduces body movements. We hypothesized that focusing on a rhythmic tactile task can anchor the child’s attention, producing a quieting effect on the motor system. The underlying hypothesis is that tactile device functions as a physical point of focus that encourages stillness. This idea is grounded in theories of embodied cognition, which suggest that calm, coordinated body states can reinforce calm internal states \citep{Dixey2024MindfulnessTA, Valentini2023YogaAM, Fang2024ASR}. 
The research extends our previous work \citep{Frederiksen2024TactileCL} that reported findings on heart rate reduction during interaction with a pocket-sized tactile device. While the earlier study focused exclusively on physiological heart rate indicators, this work presents novel findings about physical movement patterns in the same experimental context. Our previous work established that interactions with our tactile robot significantly reduced heart rate (p$<$0.01) by approximately 3-5 bpm compared to non-interaction conditions \cite{Frederiksen2024TactileCL}.
In this work, by examining physiological and behavioral dimensions together, we provide a more complete understanding of the potential of using tactile interactions to influence embodied self-regulation in children.

This work addresses the following research questions:

\begin{itemize}
    \item \textbf{RQ1:} How does tactile interaction with a pocket-sized device influence physical markers of elevated arousal in children, with a focus on movement patterns related to attention?

    \item \textbf{RQ2:} What is the relationship between physiological calming effects (measured via heart rate reduction) and behavioral regulation (measured via movement reduction) during tactile interaction with a pocket-sized device?

\end{itemize}

To answer these questions, we designed and built a simple, low-cost pocket-sized device that offers tactile vibratory interactions. By focusing on repeated button-press sequences, reinforced through gentle vibration, the device is designed to hold the user’s attention during interaction \citep{Frederiksen2024TactileCL}. Through a 14-day pilot study and a subsequent three-day in-school user study (combined \( n=20 \) participants), we gathered data from children interacting with the device, including heart-rate and video-derived joint movement tracking over time.

The contributions of this research are as follows:

\begin{enumerate}
    \item A comprehensive analysis of the collected data, showing that children experience both physiological calming (reduced heart rate) and behavioral stabilization (reduced movement) when participating in tactile interactions with our device.
    \item A demonstration of a notable calming impact on physical restlessness validated with a significant ($p<0.05$) reduction in the overall physical movement in the With-Interaction by 33.38\% compared to the Without-Interaction condition.
    \item Insights into how handheld tactile technology can serve as an accessible tool for supporting physiological calmness and self-regulation in children by simultaneously reducing physiological arousal and motor restlessness through a simple, engaging interaction that requires minimal training.
\end{enumerate}

Our findings provide an empirical foundation for understanding how tactile game engagement supports self-regulation in everyday settings. The presented study involves children without clinical diagnoses; however, the results can potentially inform future explorations of tactile technologies aimed at supporting calmness and emotional balance in broader user populations, including those experiencing anxiety. 

This paper is structured as follows: Section 2 reviews related work on assistive technologies for therapeutic applications; Section 3 describes our hypotheses, methods, and experimental design; Section 4 presents the results; Section 5 discusses those results and concludes the paper.


\subsection{Therapeutic Behavioral Interventions for Children}

Current therapeutic approaches that promote calmness include a variety of meditative practices. Targeted relaxation training has been shown to decrease motor restlessness and impulsive movement in children, fostering greater physical stability and attention span \citep{Schilling1983BehavioralRT}. Research into structured respiration practices demonstrated that brief, daily interventions can effectively reduce physiological arousal \citep{Balban2023BriefSR}, while mantra-based meditation produced effects comparable to some pharmacological interventions \citep{lvarezPrez2022EffectivenessOMq}. Self-regulation strategies like self-talk, deep breathing, progressive muscle relaxation, and imagery have been established to reduce physiological arousal. In the physical domain, structured exercise protocols  have been shown to have a positive effect on physical arousal \citep{Servant2019NonpharmacologicalTF}, with longitudinal research demonstrating improved outcomes from developmentally appropriate physical activity \citep{Melnyk2009ImprovingTM}. Experimental work has shown that even brief exercise sessions can have a positive impact on both psychological and neuro-biological processes \citep{Smits2008ReducingAS}. 
There is an extensive body of research examining the nexus between bodily focus and psychological well-being in pediatric populations. Recent evidence suggests that physiological regulation serves as a critical entry point for cognitive improvement \cite{Wang2024AerobicEP}. for instance, digital breath-focused mindfulness training in parent-child dyads has been shown to reduce default mode network activity and sharpen attentional focus in children \cite{Jaiswal2025BreathFocusedMA}. This relationship extends to broader physical engagement, where longitudinal data indicates that physical activity at age 11 acts as a significant protective factor against adolescent depression \cite{Lundgren2025ImpactOP}. Similarly, physical relaxation techniques have proven effective across diverse developmental stages and environments. Progressive muscle relaxation and similar protocols have yielded significant gains in attention and executive functioning for both kindergarten-aged children (5 to 6 years) and those in institutional settings \cite{Jarraya2022KindergartenBasedPM, Nair2024ProgressiveMR}. Building upon these somatic foundations, recent adaptations of Mindfulness-Based Cognitive Therapy (MBCT) have pivoted toward targeting emotional regulation directly \cite{Chambers2009MindfulER, Gross1998TheEF, Kuyken2022EffectivenessOU}. These interventions show promise in alleviating clinical symptoms through heightened emotional awareness \cite{Kholghi2025TheEO, Dunning2018ResearchRT}. Furthermore, the scalability of such emotional regulation support is being enhanced through repeated momentary interventions \cite{Pavlacic2024SystematicRO}, and for adolescents with ADHD, embedding these practices into daily life offers a particularly viable path for sustained behavioral health \cite{Murray2024ANR}.

While traditional mindfulness and physical activity interventions effectively promote calmness, they typically rely on adult facilitation, structured environments, and repeated training. There remains a significant need for immediate, accessible tools that allow children to engage in self-regulation autonomously within everyday contexts. This study addresses this gap by investigating whether a handheld, tactile device can provide a low-barrier, in-the-moment alternative to facilitated therapeutic practices

\subsection{Tactile Interactions and Movement Analysis}

Beyond emotional and physiological indicators, physical movements like fidgeting, pacing, and restlessness may indicated reduced self-control in children \citep{Association2022DiagnosticAS, Rapee2009AnxietyDD}. Developmental research indicates that physically active or restless behavior impedes sustained attention, hinders learning, and intensifies children's distress \citep{Diamond2014ExecutiveF, Blair2008BiologicalPI}. Interventions using mindfulness or relaxing touch demonstrate reductions in fidgeting behaviors, indicating that targeted tactile engagement as well as meditative and mindful intervention strategies may influence the relaxation state and physical movement patterns \citep{CartwrightHatton2006AnxietyIA, Silva2023SocialAA, Thompson2008MindfulnessWC, Alimardani2020RobotAssistedMP}. Previous research has indicated that engagement with structured tactile tasks produced an "anchoring" effect that reduced autonomic arousal and motor restlessness, resulting in children being calmer and less fidgety during game-like sessions \citep{MeagherPhD2001PainAE}.

While therapeutic studies rarely explicitly focus on movement, some research in the context of autism spectrum disorders has used motion-sensing technologies to correlate physical posture with emotional engagement \citep{Clabaugh2019LongTermPO, Anzalone2015EvaluatingTE}. Various studies have also used user pose-tracking to document both engagement and posture changes to provide objective measures of attention, comfort, and interaction quality in therapeutic robotics \citep{Jain2020ModelingEI, Kennedy2015TheRW}. Movement vector analysis, which tracks instantaneous body displacement, is a relatively new relaxation method that offers quantitative insights into the relationship between movement and emotion \citep{Lang2000FearAA}, an aspect that was traditionally measured using wearable devices such as heart rate and EEG monitors \citep{Michels2013ChildrensHR, Mohammadi2017WaveletbasedER}.
Recent research efforts highlight that while socially assistive robots are widely used for motivation, a critical need remains for structured tools to specifically measure the tactile dimension of these therapeutic interactions \citep{Pirborj2025TactileIW, Arnold2017TheTE, Paterson2022InvitingRT}. Parallel to this, the development of sensor-integrated fidget objects has enabled the objective quantification of restlessness, validating fidgeting patterns as measurable data points for monitoring health states \citep{Bobrova2025CipherPalAP, Bobrova2024DesignAD, Bobrova2025DesigningBW}. In the context of dementia care, interactive textile tools equipped with capacitive sensing have been shown to significantly enhance emotional engagement compared to standard non-digital sensory aids \citep{Tang2025DesignOA}. Physiological regulation has also been achieved through 'affective stroking' using mid-air haptics, where simulated tactile sensations successfully shifted autonomic states toward relaxation \citep{He2024AffectiveSD}. While significant research efforts focus on introducing Large Language Models into therapeutic scenarios, recent implementations in social robotics suggest that a robot's physical presence and gestural movement remain crucial for 'anchoring' interactions and maintaining cognitive attention \citep{Bertacchini2023ASR, Jamil2024OnCT, LeTendre2023SocialRI}. Furthermore, new methods in sensory substitution demonstrate that the emotional intent of tactile gestures can be accurately recognized through sonification, implying that touch contains extractable data features distinct from physical contact \citep{deLagarde2025PavingTW}. These findings align with theoretical frameworks that increasingly view therapeutic touch as a bidirectional communicative act, where improved 'fluency' in tactile exchange correlates directly with better functional movement outcomes \citep{Tuttle2023DevelopingFI}.

\subsection{Therapeutic Technical Behavioral Interventions for Children}

Almost thirty years of research into the use of robots with socially assistive capabilities has demonstrated a wide range of positive effects, including on physical rehabilitation for children with motor disabilities \citep{Buitrago2020AML}. Socially assistive robots (SAR) \citep{FeilSeifer2005DefiningSA} have also been shown to have significant potential for supporting children with psychological challenges by serving as child-friendly mediators facilitating gentle interactions that aid or train children to navigate social scenarios \citep{Okamura2010MedicalAH, Kozima2008PlayfulRF, Costescu2014TheEO} and provide therapeutic interactions for those experiencing cognitive deficits \citep{ShibataRegenerativeAT, Shibata2012TherapeuticSR, Tapus2009TheUO}, while employing human-robot empathy frameworks to convey comforting social cues through storytelling, emotional engagement, and therapeutic conversations \citep{Elgarf2024FosteringCC, Spitale2022SociallyAR, Lakatos2023, 10.1007/978-3-030-26945-6_1, Bgels2006FamilyCB, Leite2014EmpathicRF}. Children demonstrate high motivation and engagement when interacting with socially assistive robots (SARs) when introduced as a way to reduce fear and pain, but the current body of research remains inconclusive regarding their clinical effectiveness \cite{trostmataric2019}. However, a 2020 study found that children were able to identify a SAR designed to display empathic characteristics and reported that it assisted in managing both the pain and fear associated with IV insertion \cite{trostmataric2020}. Similar findings were outlined by \cite{Neerincx2025SociallyAR} who found that children who actively participated in interactions with a socially assistive robot, such as by performing guided breathing exercises or hugging the robot, experienced significantly greater reductions in fear and anxiety during vaccinations compared to those who only observed the robot passively.
Therapeutic robots have been deployed in diverse applications to support children, from relaxation-oriented designs like the pillow-shaped TACO robot \citep{OBrien2021ExploringTD}, and the huggable Probo robot \citep{Saldien2010ExpressingEW} to pet-inspired interactions for pain reduction such as with the Maki robot and the Paro robot seal \citep{KELLY202137}. These robots aim to build lasting relationships \citep{Daz2011BuildingUC}, and provide relaxing interactions \citep{Rabbitt2015IntegratingSA}. Robot companions specifically designed for children can reduce distress and build skills \citep{Arnold2016EmobieAR, Jeong2015ASR}, while AI-enhanced socially assistive robots offer personalized support through dynamically adjusted behaviors or empathy \citep{Trost2020SociallyAssistiveRU, Addlesee2024AMC, Addlesee2024MultipartyMC, Scassellati2012RobotsFU}. Some research emphasize the importance of personalized support, an argues that for socially assistive robots to maintain long-term therapeutic efficacy in children, they must evolve beyond static interactions to offer adaptive behaviors that sustain engagement over longer durations \cite{Rakhymbayeva2021ALE}.  These technologies address challenges by providing opportunities for social skill practice  \citep{zhu2021}, while facilitating relaxation techniques such as deep breathing and interventions that train calmness over time \citep{Matheus2023DeepBP, Matheus2022ASR, Shao2019YouAD, Crossman2018TheIO}. 

Multiple studies document tangible benefits of repeated therapeutic interactions with robots for children with physical or cognitive disabilities \citep{Alemi2016ClinicalAO, Taheri2017ClinicalIO} \citep{Marti2005EngagingWA}. In clinical settings, in particular hospitals, socially assistive robots can provide relaxation when integrated into routine care \citep{Jeong2018HuggableTI, Okita2013SelfOthersPT}. Meta-reviews show encouraging preliminary results across different healthcare contexts \citep{Littler2021ReducingNE, Logan2019SocialRF}. A longer running study from 2019 introduced a hierarchical human-robot learning framework that utilized reinforcement learning to personalize challenge levels and feedback for children with autism spectrum disorder, demonstrating that autonomous in-home robots can foster long-term cognitive gains and sustained engagement \cite{Clabaugh2019LongTermPO}.
Robot-assisted strategies employed in behavioral interventions can be diverse and can range from playful companionship to structured coaching \citep{Admoni2017SocialEG, Rasouli2022PotentialAO}. Advancements in affective computing have enabled SARs to utilize real-time facial and vocal emotion analysis to dynamically adjust their therapeutic pacing, resulting in improved emotional comfort in children during robot assisted therapy \citep{s21155166, Kim2025BiosensingDrivenCA}.

Socially assistive robots effectively motivate and aid children, but many platforms are large, cost-prohibitive, and focused on social rather than physical regulation. There is a potentially under researched area calling for of empirical data on how haptic feedback alone, isolated from social cues, influences a child's physiological and behavioral state. This research utilizes a minimalist hardware platform to evaluate whether rhythmic haptic interaction can serve as a primary mechanism for inducing embodied calmness.
\section{Materials and Methods}

\subsection{Participants} 
We recruited a total of 20 participants (N=2 for the pilot study; 2 male, and N=18 for the main study; 10 female,  8 male) aged 6 to 8 years from a single first-grade class within the Danish public school system. While the participants were classmates and familiar with one another, the study was designed for individual participation, ensuring no interaction occurred between subjects during data collection.

Recruitment was coordinated through the school's administration, starting with an agreement from the head of the school to recruit from a class of 22 students. From this initial pool, three children were excluded because their parents did not provide written consent, and one child was excluded after declining to provide verbal assent. The study protocol was approved by the Research Ethics Committee at the IT University of Copenhagen. To ensure ethical compliance, parents provided written informed consent prior to the study, and each child gave verbal assent immediately before participating. Participation was strictly voluntary, and no compensation was offered to the children or their families. Our experiment was conducted over three consecutive days, in a designated familiar and quiet classroom at the school, between 9am and 1pm. The first half of the children's school day was selected as a time of highest levels of alertness and cognitive engagement compared to the post-lunch period \cite{Latino2025SchoolBasedPA, Schrder2016LunchAS}.

\subsection{Study Design}
Our research employed a mixed-methods experimental design with the study consisting of two phases: a 14-day pilot study with two participants, followed by a main study with 18 participants. We used a within-subjects design in which each participant experienced both experimental conditions with the only difference being that the tactile interaction was enabled on the device. The conditions were:
\textbf{With-Interaction Condition}: Participants actively engaged with the robot's tactile rhythm-matching interaction by pressing its buttons in response to perceived vibration patterns.
\textbf{Without-Interaction Condition}: Participants held the same robot with the tactile game disabled, so the robot provided no vibration. This was the control condition. 
\begin{figure}[h]
\centering \includegraphics[width=0.90\textwidth]{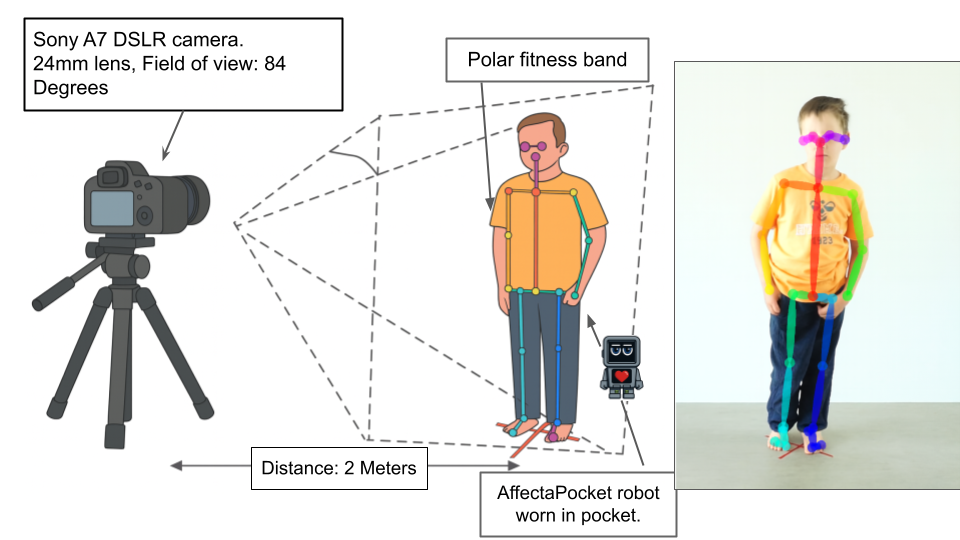} \caption{Left: The experimental setup of the main study. The participants were standing two meters away from the camera with the tactile robot in their pocket. Their heart rate were measured by a Polar heart rate tracker placed at the upper part of their dominant arm. Right: A participant interacts with the AffectaPocket in their pocket. The coloured lines are the tracked joints outputted by the open source joint pose tracking library.}
\label{experimental_setup}
\end{figure}

\subsection{Dependent Variables}
\label{sec:dependent_vars}
We measured two primary dependent variables in each experimental condition:

\begin{itemize}
\item \textbf{Heart Rate}: Continuous measurement of beats per minute (BPM) using a wrist-worn optical heart rate sensor, providing a physiological indicator of potential arousal. This Photoplethysmography-based (PPG) device was selected for its balance of accuracy and non-invasiveness, minimizing discomfort for the children during the study. Heart rate variability (HRV) rather than heart rate might be more indicative of stress, but current research suggests that measuring HRV demands extended measurement periods incompatible with our brief intervention paradigm \citep{Michels2013ChildrensHR, Shaffer2017AnOO, Malik1996HeartRV, Laborde2017HeartRV}. While optical PPG sensors are typically calibrated using adult data, the findings in this work rely on a within-subjects design where each participant serves as their own baseline. By focusing on the relative difference between the control and experimental conditions for each participant, any potential calibration offsets or absolute measurement inaccuracies are held constant across conditions.
\item \textbf{Movement Vectors}: Quantitative assessment of physical movement captured through a camera at 30 frames per second, monitoring displacement between frames as a measure of physical movement vs. calmness. Using the open-source motion analysis toolkit OpenPose (a joint-based pose tracking Python library) , each video frame was analyzed to identify the child's approximate joint locations \cite{8765346,simon2017hand,cao2017realtime,wei2016cpm}. We manually inspected data for artifacts (partial occlusions, child out of frame). Any identified joints with missing measurements were replaced with the last known joint position in the final dataset.
\end{itemize}

Together, these measures allowed us to examine both a physiological response (heart rate) and  behavioral manifestations (movement) of potential regulatory effects \citep{Healey2005DetectingSD, Sharma2012ObjectiveMS}. By capturing these variables simultaneously, we could investigate potential relationships between physiological calming and behavioral stabilization during tactile interactions. We defined attention-associated features (eyes, nose, neck, shoulders) based on the hierarchy of social attention cues \cite{Emery2000TheEH, Hietanen2002SocialAO}.In this hierarchy, head orientation can often be used instead of gaze when the eyes are not visible, while body orientation (shoulders/trunk) acts as a coarse indicator of attentional direction. Research confirms that orienting to a social stimulus typically involves a coordinated cascade where the head and torso align to stabilize the visual field \cite{doshi2012, Langton2000TheMI}. Here we isolated these features to distinguish specific orienting behaviors from general motor restlessness.
Based on the research questions outlined in Section \ref{sec:intro}, we formulated the following hypotheses:
\begin{itemize}
\item \textbf{H1}: The child participant's total physical movement will be significantly reduced during active engagement with the robot's tactile game compared to the Without-Interaction condition.
\item \textbf{H2}: The child participant's attention-displaying body parts (head, eyes, and upper body) will show greater movement reduction than other body areas during tactile interaction.
\item \textbf{H3}: The magnitude of the child participant's heart rate reduction will positively correlate with the degree of movement reduction, indicating an integrated regulatory response.
\item \textbf{H4}: The magnitude of the calming effect will be time-dependent, resulting in a more pronounced difference between the With-Interaction and Without-Interaction conditions during the final phase (the last 60 seconds) of the session compared to the initial phase.
\end{itemize}
\subsection{Handheld Tactile Robot}
\begin{figure}[]
\centering \includegraphics[width=0.80\textwidth]{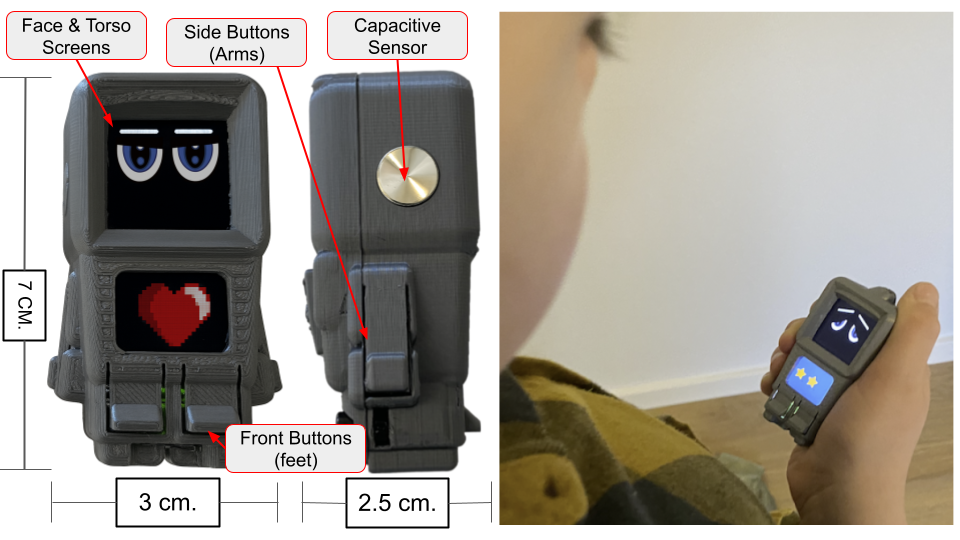} \caption{The "AffectaPocket" device, developed to promote physical and emotional self-regulation for children aged 6-12. AffectaPocket is controlled by an onboard EPS32 chip, and powered by a 3.7V battery. Children interact with the robot through a tactile rhythm-matching game played by grasping the robot in one hand and repeating generated rhythmic patterns.}
\label{affecta_annotated}
\end{figure}

\begin{figure}[h]
\centering \includegraphics[width=0.90\textwidth]{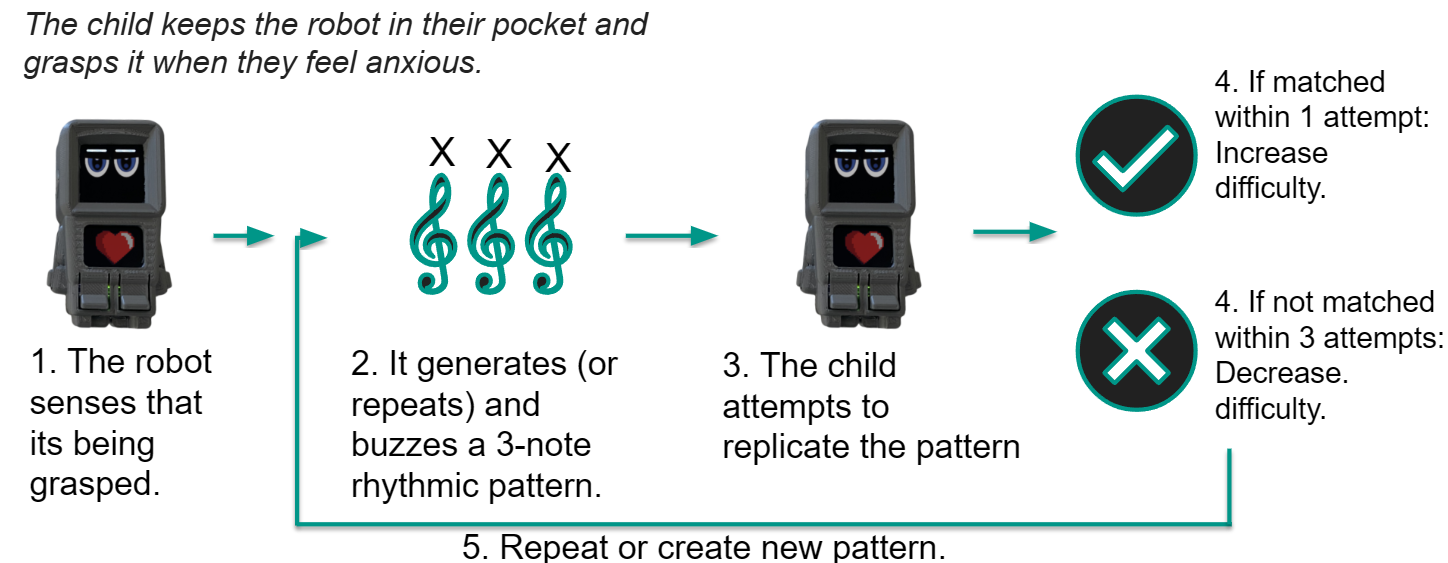} \caption{The rhythm matching interaction cycle of the AffectaPocket. The interaction is meant to occur while the robot is concealed in a pocket. The child initiates the interaction by grasping it. The robot proceeds by generating a random 3-pulse rhythm. The child attempts to replicate the rhythm. If the child succeeds in matching all pulses, a new rhythm is generated, slightly harder if the rhythm was matched in first or second attempt. The demands for precision are lowered if 3 attempts has passed with no matched rhythms.}
\label{rhythm_matching}
\end{figure}

AffectaPocket is a pocket-sized handheld device shaped like a robot that was specifically designed for children aged 6-12 to promote emotional and physical self-regulation through tactile interactions. The device can be considered a very simple robot because it has very basic intelligent control, it senses through capacitive touch and clickable buttons, and it actuates through an 8mm Linear Resonant Actuator (Model C08-005, Precision Microdrives) that provides high-fidelity haptic feedback (vibration and button click sensations) with a rated vibration frequency of 235 Hz and a normalized amplitude of 1.28 G.
The device measures 7cm × 4cm × 3cm, sized to fit comfortably in a 6-12 year old child's hand or pocket. Its anthropomorphic design in the form of a toy robot was chosen to encourage engagement and emotional connection and is based on insights from participatory design sessions with children \cite{Frederiksen2024TowardAP}. The robot has a 3D-printed gray shell with a structured body that includes articulated "arms" (side buttons) and "feet" (front buttons) that serve as clickable points. Its face consists of a small screen displaying animated eyes, while a second screen on its "torso" shows functional information and visual game cues, including a heart graphic during normal operation.
The hardware design includes the following key components (shown in Figure \ref{affecta_annotated}): capacitive touch sensors embedded in the robot's body (one circular sensor visible on the side), a battery compartment integrated into the back portion of the design, front buttons positioned as the robot's feet, side buttons that function as the robot's arms, and a USB-C charging port. The ESP-32 development board serves as the robot's brain, providing computational capabilities and Bluetooth connectivity.
The user's primary interaction with the device involves a tactile rhythm-matching game designed to hold children's attention during moments of restlessness. The game concept is outlined in Figure \ref{rhythm_matching}. When grasped for three seconds, the robot generates a three-pulse vibration pattern that the child attempts to replicate by pressing the robot's side buttons in the correct rhythm. Successful matching triggers tactile feedback through vibration and the game progresses to a new rhythmic pattern to match. This continues until the grasp on the device is released and it is left untouched for 5 seconds, at which point the pattern-matching process stops. 
The rhythm-matching game requires the user to press the side buttons simultaneously as each tactile pulse occurs, requiring the child to focus on sensing, remembering, and repeating the rhythmic pattern. A pulse is considered matched if the button press occurs within 300 ms of the pulse onset and duration. The robot repeats the full rhythm pattern, and each pulse only needs to be matched once across all attempts. After three unsuccessful attempts, the timing threshold gradually expands with 10 percent in each direction (early, late), continuing until the user just have to grasp the device to match the rhythm, to ensure eventual success. This design choice aims to prioritize user engagement and accomplishment over challenge, in order to promote positive interaction experiences for children with various ability levels.
The robot provides an initial tutorial session where a visual aid helps children to understand the interaction concept; after that, the robot is used without visual cues, allowing for discrete operation when concealed in a pocket. This design choice specifically addresses the need of children to play privately and not draw attention to self-regulating strategies.
\subsection{Pilot Study: Initial Heart Rate Assessment}
We conducted a preliminary 14-day pilot study with two typically developing children (aged 8 years) to evaluate AffectaPocket's impact on heart rate. Testing was conducted in participants' homes at consistent times (3-4pm daily), with each interaction lasting one minute. Participants completed the two study conditions daily, in randomized order (determined by a coin toss), separated by a 20-minute wash-out interval during which the the participants left the device on a nearby table. During the study conditions, the robot remained in participants' pockets. The participants were free to choose which pocket and which hand they used when interacting with the device. They put their hand in the pocket during the sessions in both conditions, ensuring that only tactile stimulation differentiated the two conditions. If at any time the children wore outfits with no pockets, they were asked to hold the device behind their backs.

\subsection{Procedure}
The study procedure was as follows:

\begin{enumerate}
\item Each participant was fitted (by the experimenter) with the heart rate tracking band on their upper non-dominant arm.
\item The participant was given a short introduction by the experimenter on how to interact with the robot. This included how to grasp the robot and how to match the rhythm of the vibrational pulses.
\item The participant was asked by the experimenter to stand in the middle of the room, on a spot on the ground marked with red tape.
\item The same room was used for all participants throughout the two weeks of the experiment.
\item Heart rate tracking was initiated.
\item The participant then completed the following:
\begin{itemize}
\item Condition 1. \textbf{\textit{With-Interaction}}: initiate the tactile interaction game by grasping the robot's side buttons for three seconds.
 \item Condition 2. \textbf{\textit{Without-Interaction}}: Similar to the With-Interaction conditions with the difference being that the robot would not have the tactile interactions enabled.
 \end{itemize}
\item A timed 20 minute break was held in between the two conditions.
\end{enumerate}

The experimental setup is illustrated in Figure \ref{experimental_setup}. For the Without-Interaction condition, the child was allowed natural movement (they were not given instructions to stand or keep still). This methodological choice was based on research showing that children have limited attention spans (chronological age plus 1 minute) and that fidgeting represents a normal self-regulatory mechanism for attention and stress \citep{Farley2013EverydayAA, Swartz2020AttentivenessAF}. Requiring sustained stillness could introduce artificial constraints that could impact the measurement of natural movement differences between conditions.

\subsection{Main Study: Heart Rate and Physical Movement Measures}
We used the same procedure as in the pilot study. Each session began with the child completing a tutorial interaction on the robot before completing both experimental conditions in randomized order. A 10-point Likert scale questionnaire (in Smiley Face Likert format) was administered by the experimenter after the interactions to assess the child's self-reported calmness and beliefs about the robot's ability to provide emotional support.

\subsubsection{Participants and Sample Size Determination}
We conducted a power analysis to determine the appropriate sample size for our main study based on data from our 14-day pilot study, as shown in Table \ref{mean_rates}. The power analysis values produced a pooled standard deviation of approximately 8.1 bpm and an effect size (Cohen's d) of 0.77. With a significance level of a=0.05, and a desired power of 80\% for a two-tailed test using a paired design, the power analysis indicated a minimum required sample size of 14 participants. To account for potential variability in a larger sample and possible minor data loss, we increased our target by 30\%, resulting in our final sample of 18 children in the main study.

\begin{table}[h!]
\centering
\begin{tabular}{lccc}
\hline
\textbf{Participant} & \textbf{With Interaction (bpm)} & \textbf{No Interaction (bpm)} & \textbf{Difference (bpm)} \\
\hline
A & 93.1 & 99.5 & 6.4 \\
B & 96.4 & 102.4 & 6.0 \\
\hline
\textbf{Overall Mean} & 95.0 & 101.0 & 6.0 \\
\textbf{Std. Deviation} & 7.0 & 9.0 & -- \\
\hline
\end{tabular}
\caption{Mean heart rates (in bpm) for participants with and without interaction.}
\label{mean_rates}
\end{table}

\subsection{Data Processing and Statistical Methods}
As introduced in Section \ref{sec:dependent_vars}, we approximated body stillness by quantifying the relative displacement of tracked body joints over time. This approach transformed raw video data into a sensitive behavioral correlate for restlessness, allowing for a continuous movement metric that could be statistically compared against heart rate fluctuations and self-reported measures. 
For each captured body joint \( j \) and image frame \( f \), the movement distance \( D_{j,f} \) was calculated using standard Euclidean distance (in image pixels) between consecutive frames, measured as:
\begin{equation}
D_{j,f} = \frac{\sqrt{(x_{j,f} - x_{j,f-1})^2 + (y_{j,f} - y_{j,f-1})^2}}{\Delta t}
\end{equation}
where \( (x_{j,f}, y_{j,f}) \) are the pixel coordinates of joint \( j \) in frame \( f \), and \( \Delta t = \frac{1}{30} \) seconds is the frame interval.
The lengths of these movement vectors were transformed to SI standard measurement of meters based on the cameras field of view (84 degrees) and distance to the tracked objects (2 meters). Total joint movement was computed as the sum of these distances per second. While pixel-to-distance conversion introduces measurement error, the error was systematic and affected both conditions equally, so it did not bias the comparisons between conditions.
Statistical analysis consisted of paired-samples t-tests to assess within-subject differences in both heart rate and movement between conditions. For heart rate analysis, we examined the mean differences in beats per minute between conditions. For movement analysis, we compared average displacement per second between conditions. We also calculated Pearson's correlation coefficient to explore relationships between heart rate reduction and movement changes, to explore whether physiological calming corresponded with behavioral stabilization. Effect sizes for significant differences were calculated using Cohen's d. 
We also examined correlations between physiological responses and self-reported attitudes using Pearson's correlation coefficient (r) as well as using the Random Forest algorithm. Heart rate (bpm) and physical movement (mean displacement across 15 body points) were correlated with Likert questionnaire responses to assess self-reported calmness and beliefs about robot helpfulness. 

\section{Results}

\subsection{Heart Rate}
\begin{figure}[ht]
\centering
\includegraphics[width=1.00\textwidth]{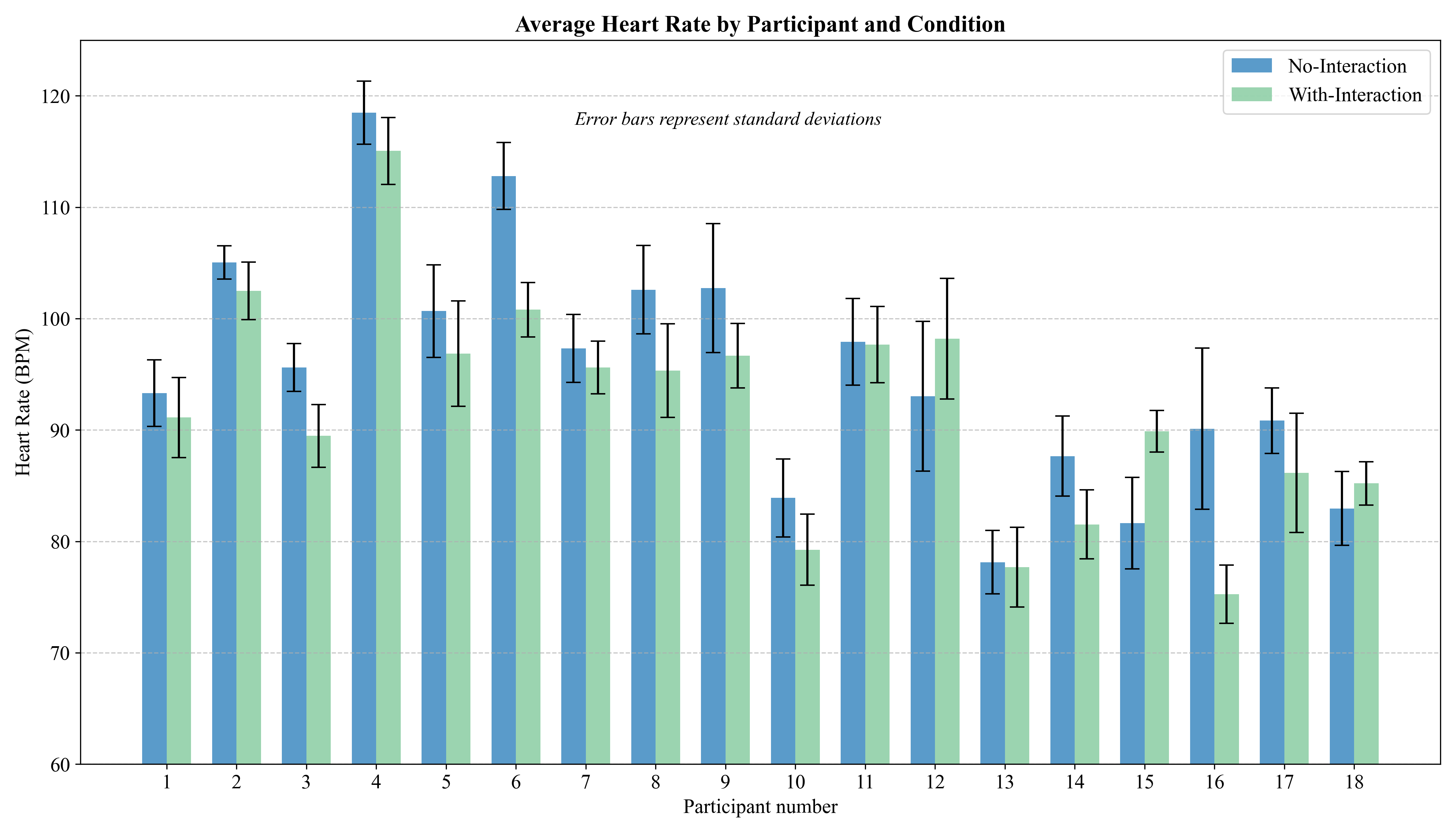}
\caption{The average difference in heart rate for each of the two conditions per participant; blue bars indicate the Without-Interaction condition and green bars the With-Interaction condition. (Figure reproduced from results in \citep{Frederiksen2024TactileCL}).}
\label{old_heart_rate_results}
\end{figure}

Figure \ref{old_heart_rate_results} presents the average heart rate measurements for each participant for both experimental conditions. On average, participants experienced a heart rate reduction of 3.37 bpm during the With-Interaction condition compared to the Without-Interaction condition, with a standard deviation of 7.32 bpm. A paired-samples t-test demonstrated a significant difference between conditions ($p<.01$), confirming that active tactile interaction with the robot produced consistently lower heart rates than only holding the passive device in the control condition. The effect size (Cohen's d = 0.32) indicates a small-to-medium effect.
Notably, the 3.37 bpm average reduction is comparable to heart rate decreases achieved through established relaxation techniques such as meditation or controlled breathing exercises, which typically require daily extended practice periods of 20 minutes or more to be effective \citep{Russo2017ThePE, Zaccaro2018HowBC, Perciavalle2017TheRO}. The brief interaction with AffectaPocket achieved similar physiological calming effects within two minutes and without prior training.

\begin{figure}[h]
\centering
\includegraphics[width=1.0\textwidth]{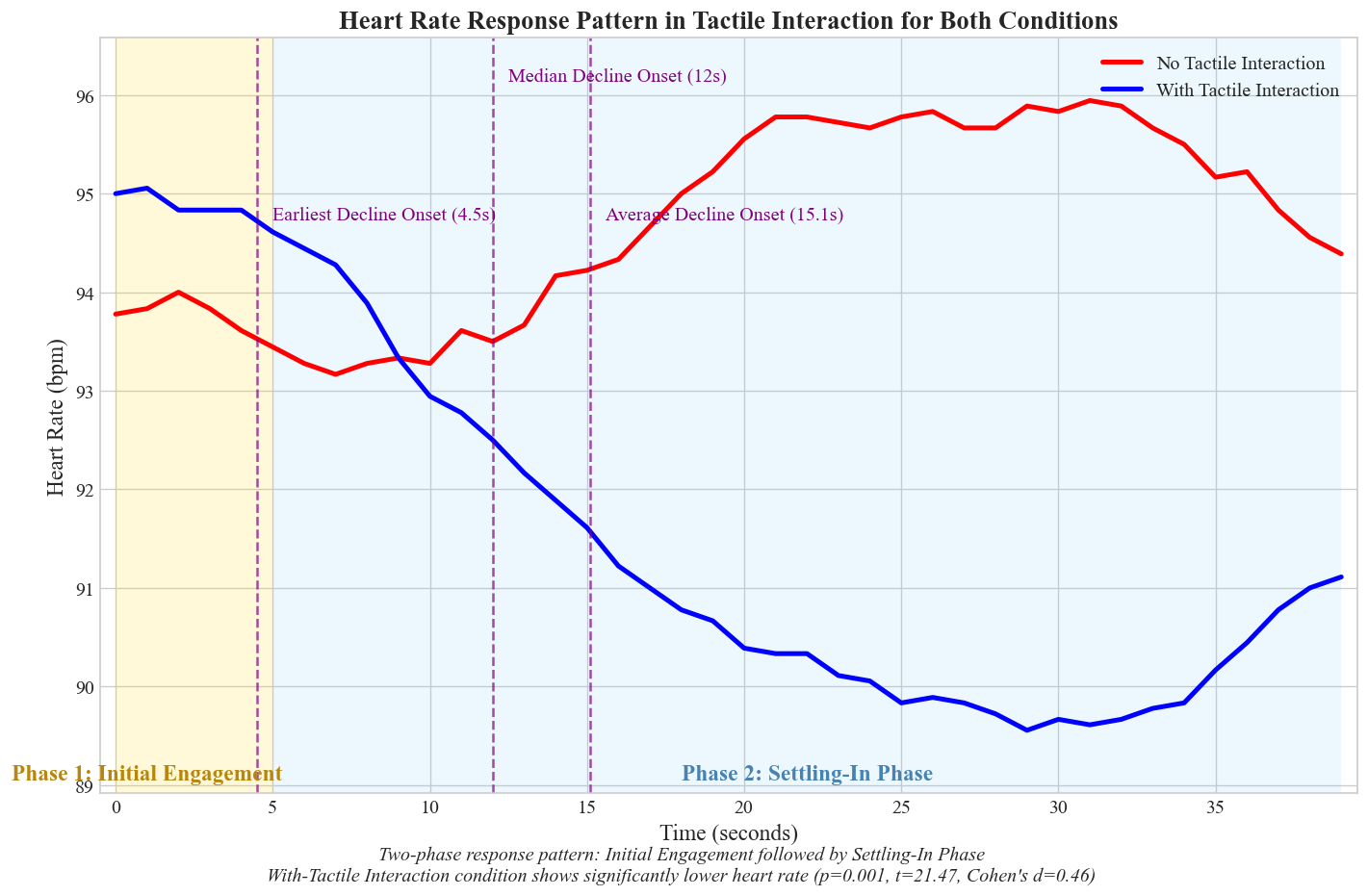}
\caption{The average difference in initial onset times for heart rate changes for each of the conditions over the initial 40 seconds of the interaction. The blue line shows the With-Interaction condition and the red line shows the Without-Interaction condition.}
\label{fig:heart_rate_onset}
\end{figure}

An analysis of the onset timing of changes in heart rate during the interaction with the handheld robot revealed a two-phase response pattern in the With-Interaction condition, as outlined in Figure \ref{fig:heart_rate_onset}. During the Initial Engagement Phase (first ~5 seconds), 44.4\% of participants (8 out of 18) showed a clear heart rate pattern where their measurements were initially stable or slightly elevated in the With-Interaction condition. 
At the first recorded measurement point (time = 0s), the With-Interaction average was 95.0 bpm (SD=11.26), whereas the Without-Interaction condition was 93.78 bpm (SD=10.43). This could represent initial engagement with the tactile stimulus. Following the first 5 seconds, a Settling-In Phase began, with heart rate declining as participants become more focused. This decline seemed most pronounced in the largest cluster of participants (50\% of those showing a pattern of declining heart rate), beginning between 3 to 6 seconds (average 4.5 seconds), while others showed similar patterns with delayed onset times (median onset at 12 seconds). The overall lower heart rate in the With-Interaction condition (M = 92.33, SD = 10.43) compared to Without-Interaction condition (M = 95.73, SD = 11.26) appears to be highly impacted by this early settling-in effect, as shown in Figure \ref{fig:heart_rate_onset}.  Focused engagement with the tactile interaction leads to a calmer physiological state that persists throughout the session, as demonstrated by 83.3\% of participants (15 out of 18) showing lower average heart rates in the With-Interaction condition.
\subsection{Body Movement}
\label{sec:overal_movement}
\begin{figure}[h]
\centering
\includegraphics[width=1.00\textwidth]{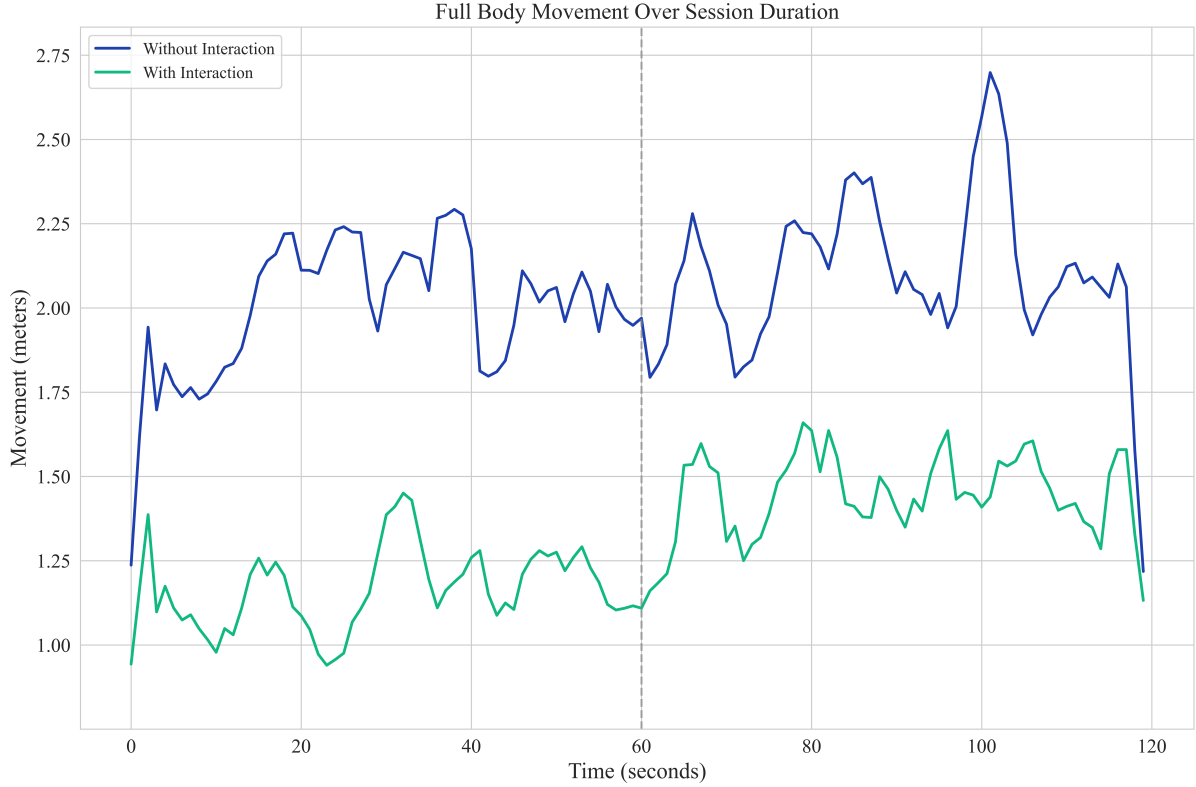}
\caption{This time series depicts participant-level average body movement vectors including all tracked joints for each condition With-Interaction and Without-Interaction across the two minute study. The graph has been smoothed with rolling average. Lower values indicate less movement.}
\label{full_body_time_series}
\end{figure}

The child participants in the study showed less overall movement when interacting with the tactile game in the With-Interaction condition. Table \ref{tab:movement-summary} depicts the average body movement decrease for each condition. Overall movement reduction is evident as most participants (15 out of 18, or 83.3\%) showed reduced full body movement during the With-Interaction compared to Without-Interaction, with a significant (paired-samples t-test, p$<$.01) average decrease from 223.5 cm (SD = 135.6 cm) to 139.4 cm (SD = 56.9 cm) (37.6\% reduction). Individual variability is reflected in the high standard deviations (135.6 cm for Without-Interaction, 56.9 cm for With-Interaction), showing large individual differences in the individual movement levels and responses to the tactile game intervention, with some children being more active than others during both conditions. A few children (3 out of 18) had the opposite effect: they displayed increased body movement in the With-Interaction condition (the largest increase was 31\%). Figure \ref{full_body_time_series} shows participant movement over time, depicting summed movement vectors (in meters) throughout the interaction. The plotted time series reveals that the with-tactile-interaction condition generally demonstrated lesser overall movement than the Without-Interaction condition. The consistently lower movement suggests that the tactile game may help to reduce physical restlessness.
\begin{figure}[h]
\centering
\includegraphics[width=1.0\textwidth]{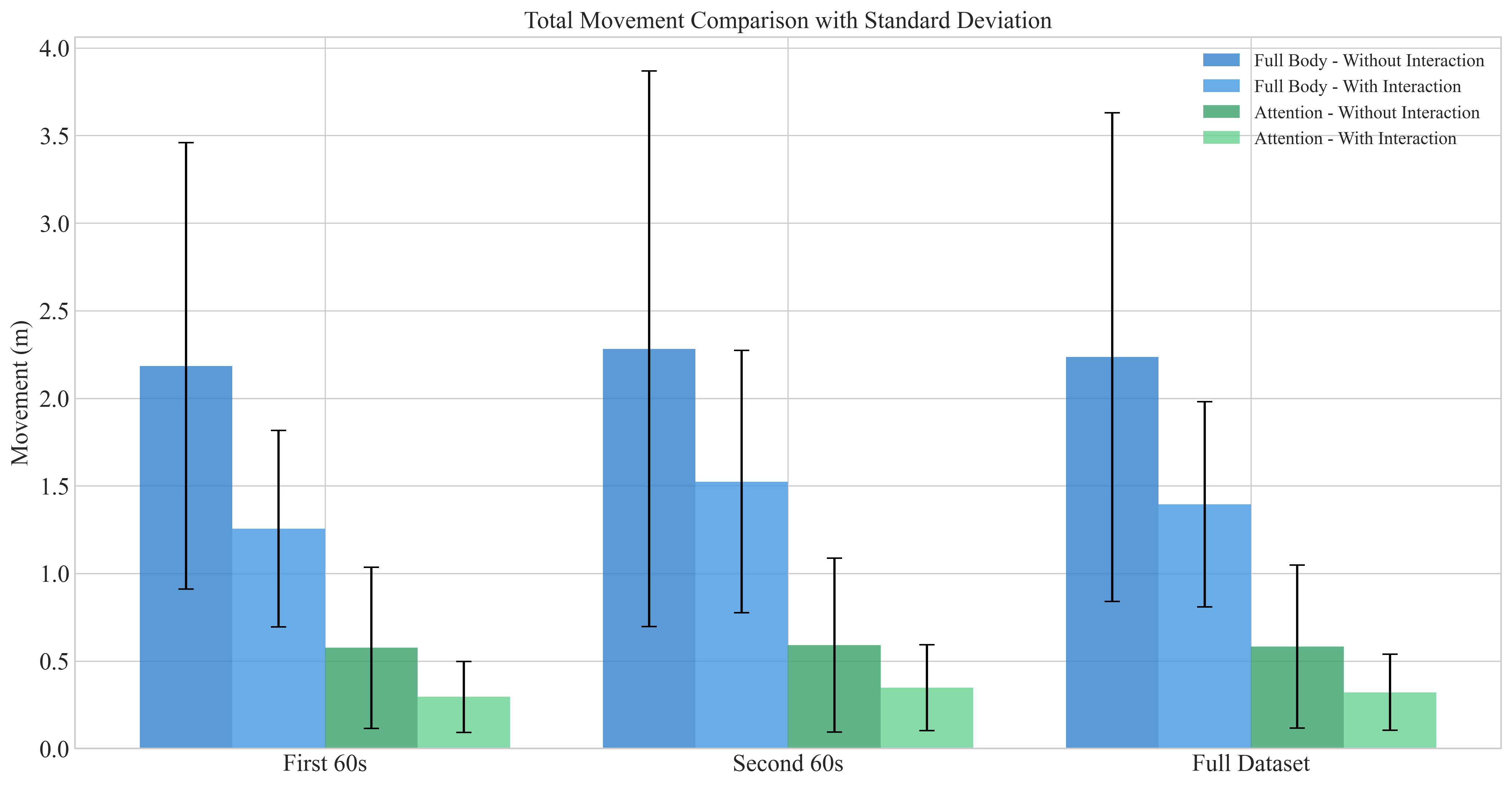}
\caption{The average total movement and standard deviation for all users per condition. The graph shows full body movements (all tracked joints, analyzed in \ref{sec:overal_movement}) and the joints associated with user-attention (analyzed in \ref{sec:attention_movement}).}
\label{fig:movement_comparison}
\end{figure}

\label{sec:attention_movement}
We compared the difference in movement of the attention-focused joints:
\begin{itemize}
\item Eyes (left and right eye), Face (nose) - related to head orientation
\item Neck - related to head movement
\item Shoulders (left and right) - related to upper body posture
\end{itemize}
When focusing only on attention-related joints, the effect of tactile interaction is more pronounced. Figure \ref{fig:movement_comparison} depicts the difference between conditions for both full body movements and attention-focused joints, highlighting the difference in calculated means between the initial 60 seconds of the interaction, the final 60 seconds of the interaction, and the full interaction. The mean movement for attention-focused joints is:
58.2 cm (SD = 45.2 cm) in the Without-Interaction condition, and 32.1 cm (SD = 21.2 cm) in the With-Interaction condition. This represents a significant (paired-samples t-test, $p<.01$) 44.\% reduction in movement during the tactile interaction, compared to the 37.6\% full body movement reduction, suggesting that tactile interaction with the robot appears to particularly affect the stability of attention-related body parts. The fact that head, eye, and shoulder movements are more markedly reduced during tactile interaction suggests that participants may be adopting a more focused or attentive pose when engaging in tactile interaction with the robot.

\subsection{Statistical and Effect Size Analyses}
We conducted paired \textit{t}-tests comparing the Without- and With-Interaction conditions for each child. The overall reduction in movement with tactile interaction was statistically significant ($p = 0.005$). By body region, tactile interaction significantly reduced total movement in the head region ($p = 0.003$), upper limbs ($p = 0.01$), and torso ($p = 0.005$). In the lower limbs, however, the difference was not statistically significant ($p = 0.23$). At the individual joint level, most joints in the head, arms, and torso also showed significant decreases ($p < 0.05$), while right hip ($p = 0.08$), right knee ($p = 0.82$), and left knee ($p = 0.15$) did not show significant decreases.

We also analyzed the magnitude of the movement difference using effect size measures. The Cohen's d for the paired difference in movement (Without-Interaction vs. With-Interaction) was calculated at d = 0.91, meaning the difference in movement between conditions is not only statistically significant but was substantial in size. The large effect indicates that tactile interaction with the robot had a substantial impact on reducing physical activity levels. 

Using an ANOVA statistical test, we found a significant main effect when comparing the conditions (F(1,17) = 7.62, $p < .05$), with a partial eta-squared of 0.31. This suggests that approximately 31\% of the variance in movement can be attributed specifically to the presence or absence of tactile interaction with the robot. Overall, the effect sizes suggests that the condition difference is not only statistically reliable but also practically meaningful.

\subsection{Temporal Interaction Analysis}
As the heart rate data suggest an increased impact within the first half of the interaction, we analyzed the movement data to compare differences in the two halves of the interaction. 

Table \ref{tab:movement-summary} presents the movement averages, comparing the initial 60 seconds against the full interaction period. The results indicate that the effect on attention-related joints was particularly robust in the initial phase, with 88.9\% (16 of 18) of participants showing reduced movement. In terms of magnitude, the movement reduction appeared numerically larger during the first 60 seconds compared to the second 60 seconds for both attention-focused joints (Mean Difference = 28.0 cm vs. 24.2 cm) and full-body movement (Mean Difference = 92.8 cm vs. 75.6 cm). However, paired samples t-tests revealed that these differences between the initial and final phases were not statistically significant (p=0.48 and p=0.40). The lack of statistical significance between phases can be explainedby high inter-participant variability (SD=83.9 cm). While 66.7\% of participants (12 of 18) showed the expected stronger effect in the first minute, some outliers exhibited a delayed response, with one participant showing a 184 cm greater reduction in the second phase. This suggests that while immediate calming is typical, individual responses vary significantly, requiring a larger sample size to appear statistically.
\begin{table}[h]
\centering
\caption{Summary of movement data comparing the first 60 seconds to the full interaction. For each measurement type, the table shows mean movement values for the Without-Interaction and With-Interaction conditions, the mean difference between conditions, and the number of participants showing reduced movement during With-Interaction.}
\label{tab:movement-summary}
\begin{tabular}{lcccc}
\toprule
\textbf{Measurement} & \textbf{Without} & \textbf{With} & \textbf{Difference} & \textbf{Participants} \\
\textbf{Condition} & \textbf{Interaction} & \textbf{Interaction} & & \textbf{Showing Effect} \\
\midrule
\multicolumn{5}{l}{\textit{First 60 Seconds}} \\
Full Body & 873.3 & 502.3 & 371.0 & 15 of 18 \\
Attention Joints & 230.1 & 118.0 & 112.1 & 16 of 18 \\
\midrule
\multicolumn{5}{l}{\textit{Full Dataset}} \\
Full Body & 893.5 & 557.5 & 336.0 & 15 of 18 \\
Attention Joints & 232.8 & 128.5 & 104.3 & 16 of 18 \\
\bottomrule
\end{tabular}
\end{table}

\subsection{Hart Rate and Self-Assessment Questionnaire Data}

Analysis of correlations between the self-reported questionnaire data and heart rate measurements revealed several relationships. A significant but small correlation was observed between children's belief in the robot's ability to help them in a stressful situation, and their heart rate changes during interaction (r = -0.34, p $<$ 0.05), suggesting that children with higher confidence in the robot's supportive capabilities exhibited less pronounced decreases in heart rate. This finding appears counter-intuitive but may be partly explained by the participants' starting states. Participants with high confidence in the robot ($\ge$8) tended to begin the session with lower heart rates (93.1 bpm, at Time=0) in comparison with those with medium to low confidence ($<$8) who started with higher heart rates (95.0 bpm, and Time=0). However, this correlation between the intial heart rate and confidence in the robot was not significant, so we cannot definitively determine whether lower heart rate was a result of anticipatory trust in the robot or attributable to other factors. A similar relationship was found between belief in the robot's effectiveness for offering support and movement differences (r = 0.23, p = 0.09), though this correlation did not reach statistical significance. Self-reported calmness showed a weak positive correlation with heart rate changes (r = 0.21, p = 0.11), suggesting that children who described themselves as calmer demonstrated smaller reductions in heart rate during robot interaction. Overall, these weak correlations require further studies to explore how preexisting attitudes about robots may influence children's physiological responses during interaction.

\subsection{Heart Rate and Physical Movement}

The analysis of the relationship between physical movement and heart rate across conditions revealed no significant correlations when examining simultaneous movement and heart rate patterns. However, when investigating time-delayed effects (first 30 seconds of movement and last 20 seconds of measured average heart rate), we found a moderate negative correlation (r = -0.37, p = 0.41), indicating that higher initial movement might predict lower subsequent heart rate. This follows intuition that the more initial movement could indicate an opportunity for  a larger heart rate drop effect, though this effect did not reach statistical significance.

\subsection{Random Forest analysis}

To explore potential non-linear relationships and interactions between the subjective variables of the post-experiment questionnaire and the heart rate and movement measurements, we analyzed the data using a Random Forest algorithm with resampling across 1,000 iterations. This analysis revealed that children's belief in the robot's ability to help was consistently the most important predictor for both physiological measures, showing moderate importance for heart rate changes (importance score: 0.34) and movement differences (importance score: 0.23). However, none of these predictors reached statistical significance. The limited sample size (n=18) of our study could explain the lack of statistical power in these results.

\subsection{Hypothesis Testing}

Our findings provide the following evidence relative to our study hypotheses:

\begin{itemize}

\item \textbf{H1: The child participant's total physical movement will be significantly reduced during active engagement with the robot's tactile game compared to the Without-Interaction condition.} \\
\textit{H1 is supported.} The overall reduction in movement for the With-Interaction condition was statistically significant (p = 0.005), with 83.3\% of participants (15 out of 18) showing reduced full body movement during the session for this condition. On average, movement decreased from 2.23 (SD = 1.36) meters to 1.39 (SD = 0.57) meters, representing a 37.6\% reduction. The Cohen's d value of 0.91 and partial eta-squared of 0.31 from the ANOVA indicate that this effect is not only statistically significant but also substantial in magnitude.
\item \textbf{H2: The child participant's attention-displaying body parts (head, eyes, and upper body) will show greater movement reduction than other body areas during tactile interaction.}\\
\textit{H2 is supported.} When focusing only on attention-related joints, the effect of tactile interaction was more pronounced. The mean movement for attention joints showed a 44.8\% reduction during the With-Interaction condition (from 0.58 M to 0.32 M), compared to the 37.6\% reduction in overall body movement. This suggests that tactile interaction with the robot particularly affects the stability of attention-related body parts.
\item \textbf{H3: The magnitude of the child participant's heart rate reduction will positively correlate with the degree of movement reduction, indicating an integrated regulatory response.} \\
\textit{H3 is not supported.} Our analysis revealed no significant correlations when examining simultaneous movement and heart rate patterns. When investigating time-delayed effects (first 30 seconds of movement and last 20 seconds of measured average heart rate), we found a moderate negative correlation (r = -0.37, p = 0.41), suggesting that higher initial movement might predict lower subsequent heart rate. However, this correlation did not reach statistical significance, and the direction was opposite to our hypothesis. This indicates that heart rate and movement reductions, while both significant individually, likely represent separate regulatory mechanisms rather than an integrated response.

\item \textbf{H4: The magnitude of the calming effect will be time-dependent, resulting in a more pronounced difference between the With-Interaction and Without-Interaction conditions during the final phase (last 60 seconds) of the session compared to the initial phase.} \\
\textit{H4 is not supported.} Our results indicated that the calming effect was most pronounced during the first 60 seconds compared to both the second half as well as the full dataset duration, with a mean difference of 0.93 Meters during the first 60 seconds versus 0.84 Meters across the full dataset (10.4\% stronger effect during the initial period). The effect on attention-related joints was particularly robust in this initial phase, with 88.9\% of participants showing reduced movement in these regions. These findings suggest that the calming effect of tactile interaction may be most potent during the early engagement period, potentially before participants become habituated to the experimental condition or before fatigue becomes relevant.
\end{itemize}

In summary, our results support two of our hypotheses (H1 and H2) regarding the calming effects of tactile interaction on physiological measures, but do not support the hypotheses regarding the correlation between movement and impact on heart rate (H3) or the time-dependent nature of changes in heart rate and movement (H4).
\section{Discussion}\label{sec:discussion}

\subsection{Supporting Physical Calmness}

The results of our study suggest that a tactile game interaction leads to a \emph{calming effect} on participanting children's physical movement. Overall activity was reduced by about 37.6\% when tactile stimulation was present, and the largest decrements appeared in the head and arm regions, implying increased stillness and possibly enhanced focus. Taken together, these results suggests the impact of tactile human–robot interaction on human physical activity: when participants are engaging in touch with the robot, they tend to move less overall, possibly indicating more focused, calm, or restricted movement behavior during those interactions. While our study focused on typically developing children, these findings align with sensory integration frameworks which propose that gentle tactile input can help soothe hyperactive behaviors, enabling children to remain more settled and attentive \citep{Ayres2005SensoryIA}. This perspective is further supported by research in Attention-Deficit Hyperactivity Disorder, demonstrating that targeted tactile interventions can significantly improve self-regulation and reduce hyperkinetic behaviors in children with attention difficulties. \citep{VandenBerg2001TheUO, Pfeiffer2008EffectivenessOD}. In theoretical terms, tactile stimulation can potentially help children achieve a closer to optimal level of arousal, thereby reducing extraneous movement. This warrants further investigations to establish if such a state of arousal is findable and achievable using tactile technology. 
Our study demonstrates that our pocket-sized tactile robots can produce immediate calming effects during brief interactions, as measured by reduced heart rate and decreased motor restlessness. However, we must distinguish between these short-term physiological and behavioral changes and long-term impact, which would require persistent effects across multiple contexts and extended time periods. The current findings support the potential of such devices as momentary intervention tools rather than comprehensive long-term tools. Longitudinal studies are essential for determining whether these immediate effects translate to meaningful long-term benefits for self-regulating restlessness and countering unwanted physical movements.

\subsection{Heart Rate and Movement Disparities}

The lack of correlation between movement reduction and heart rate changes suggests that the calming effect on heart rate may not be primarily a result of reduced physical activity. This finding indicates that the robot's tactile interaction might engage children's attention in a way that affects physiological arousal through attention regulation mechanisms that operate somewhat independently from motor behavior \citep{kendall2006cognitive, Silva2023SocialAA}. Research has shown that focused attention activities can influence autonomic nervous system activity even without significant changes in physical movement \citep{Perciavalle2017TheRO, Russo2017ThePE}. Our findings suggest that the pocket-sized robot could simultaneously influence multiple aspects of self-regulation, affecting both motor control and physiological arousal, but through different pathways. This dual effect could potentially be beneficial for a future use in anxiety management, as anxiety often manifests through both physical restlessness and elevated physiological arousal \citep{Ginsburg2006SomaticSI, Jarrett2008ACR}. 
However, the lack of correlation between movement reduction and heart rate changes warrants consideration of alternative explanations for our findings. While we propose that the tactile interaction directly influences self-regulation mechanisms, other factors could contribute to the observed effects. These include the novelty of interacting with a robot, general distraction effects that might be achievable through other engaging activities, or the structured nature of the rhythm-matching task independent of its tactile components. Future studies employing additional control conditions (e.g., expanded non-tactile game interactions) could help to further uncover the specific mechanisms underlying these effects.

\subsection{Limitations and Methodological Considerations}
The current study has several limitations. First, the small sample size (n=18) limits statistical power and generalizability of the findings. Second, the experimental setting required children to stand on a marked spot to ensure that they were captured by the camera, which introduced an artificial constraint on natural movement. This experimental constraint of being stationary in a specific location could affect children's natural movement patterns, possibly increasing restlessness in the Without-Interaction condition as children attempted to comply with instructions to remain stationary. Future studies should include a more diverse sample in less constrained environments, allowing for more naturalistic movement assessment, to determine if the effects observed here generalize to a standard school contexts in a larger sample of participating children.

\subsection{Beyond Traditional Approaches: The Role of Tactile Robots in Anxiety Intervention}
The findings from this study indicate that handheld pocket-sized tactile devices could potentially be used to address some of the physiological and behavioral markers commonly associated with anxiety through a simple, engaging play interaction. AffectaPocket simultaneously reduced both physiological arousal (avg. heart rate decreased by 3.37 bpm, $p<0.01$) and motor restlessness (overall movement reduced by 37.6\%, $p<0.05$) during brief tactile interactions. This dual-pathway intervention can potentially be particularly valuable since anxiety in children typically manifests through both internal physiological symptoms and external behavioral signs \citep{Association2022DiagnosticAS, beesdo2009anxiety, Ginsburg2006SomaticSI}. Unlike traditional therapeutic approaches that often require extensive training sessions before children can independently implement them \citep{kendall2006cognitive, Oar2017AdaptingCB, Rapee2009AnxietyDD}, our tactile handheld robot may provide more immediate benefits with little if any instruction. This suggests a novel approach in which affordable and accesible handheld devices like AffectaPocket could potentially serve as complementary tools for established therapeutic interventions by offering accessible, in-the-moment support during stressful situations when trained strategies might be difficult to recall or implement \citep{Creswell2014AssessmentAM, Velting2004UpdateOA}. Such tools are needed given the high prevalence of childhood anxiety \citep{Bitsko2022MentalHS, Widiger2022TheDA}, and could potentially help to bridge treatment gaps for children experiencing anxiety symptoms who have not yet received formal diagnosis or intervention. 
Our focus on typically developing children offers insights particularly relevant to the significant population of children experiencing anxiety symptoms while awaiting formal diagnosis or intervention. With increasing prevalence of childhood anxiety and limited access to timely mental health resources, approaches that provide immediate, accessible support could be especially valuable during this diagnostic waiting period when few targeted interventions are available. While our findings show promising effects on physiological and behavioral indicators in non-diagnosed children, extrapolation to clinical populations is not advised. instead, the current study serves as a proof-of-concept that tactile handheld robot interactions can influence markers associated with anxiety, but controlled clinical trials with diagnosed populations are necessary to establish therapeutic efficacy, and this work aims to inform such projects.


\bibliographystyle{Frontiers-Harvard}
\bibliography{bibliography}

\end{document}